\documentclass{article}

\PassOptionsToPackage{numbers, compress}{natbib}
\usepackage[preprint]{nips_2018}

\usepackage[utf8]{inputenc} % allow utf-8 input
\usepackage[T1]{fontenc}    % use 8-bit T1 fonts
\usepackage{hyperref}       % hyperlinks
\usepackage{url}            % simple URL typesetting
\usepackage{booktabs}       % professional-quality tables
\usepackage{amsfonts}       % blackboard math symbols
\usepackage{nicefrac}       % compact symbols for 1/2, etc.
\usepackage{microtype}      % microtypography

\usepackage{graphicx}
\usepackage{comment}
\usepackage{floatrow}
\usepackage{amsmath}
\usepackage{extarrows}
\newfloatcommand{capbtabbox}{table}[][\FBwidth]
\usepackage{enumitem}

\title{Unsupervised Learning of Neural Networks to Explain Neural Networks}
\author{Quanshi Zhang$^{b,a}$, Yu Yang$^{a}$, Yuchen Liu$^{c}$, Ying Nian Wu$^{a}$, and Song-Chun Zhu$^{a}$\\
$^{a}$University of California, Los Angeles\qquad $^{b}$Shanghai Jiao Tong University\\$^{c}$University of California, Irvine.
}

\begin{document}

\maketitle

\begin{abstract}
This paper presents an unsupervised method to learn a neural network, namely an \textit{explainer}, to interpret a pre-trained convolutional neural network (CNN), \emph{i.e.} explaining knowledge representations hidden in middle conv-layers of the CNN. Given feature maps of a certain conv-layer of the CNN, the explainer performs like an auto-encoder, which first disentangles the feature maps into object-part features and then inverts object-part features back to features of higher conv-layers of the CNN. More specifically, the explainer contains interpretable conv-layers, where each filter disentangles the representation of a specific object part from chaotic input feature maps. As a paraphrase of CNN features, the disentangled representations of object parts help people understand the logic inside the CNN. We also learn the explainer to use object-part features to reconstruct features of higher CNN layers, in order to minimize loss of information during the feature disentanglement. More crucially, we learn the explainer via network distillation without using any annotations of sample labels, object parts, or textures for supervision. We have applied our method to different types of CNNs for evaluation, and explainers have significantly boosted the interpretability of CNN features.
\end{abstract}

\section{Introduction}

Convolutional neural networks (CNNs)~\cite{CNN,CNNImageNet,ResNet} have achieved superior performance in many visual tasks, such as object classification and detection. In contrast to the significant discrimination power, the model interpretability has always been an Achilles' heel of deep neural networks.

In this paper, we aim to boost the interpretability of feature maps of a middle conv-layer in a CNN. \textit{Without additional human supervision, can we automatically disentangle human-interpretable features from chaotic middle-layer feature maps?} For example, we attempt to force each channel of the disentangled feature map to represent a certain object part. Our task of improving network interpretability by exploring more interpretable middle-layer features is different from the passive visualization~\cite{CNNVisualization_1,CNNVisualization_2,CNNVisualization_3,FeaVisual,visualCNN_grad,visualCNN_grad_2} and diagnosis~\cite{Interpretability,CNNInfluence,trust} of CNN representations.

A major issue with network interpretability is that high interpretability is not necessarily equivalent to, and sometimes conflicts with a high discrimination power~\cite{Interpretability}. For example, disentangling middle-layer representations of a CNN into object parts~\cite{interpretableCNN} may sometimes decrease the classification performance. Thus, people usually have to trade off between the network interpretability and the network performance in real applications.

\begin{figure}[t]
\centering
\includegraphics[width=0.99\linewidth]{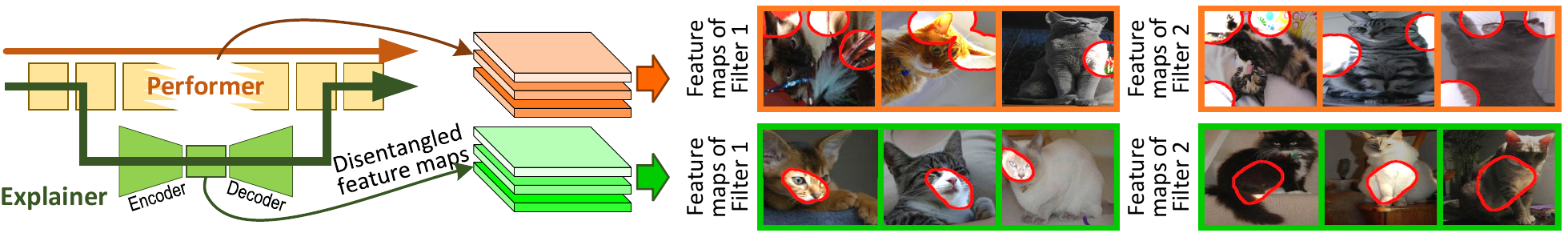}
\caption{Explainer network. We use an explainer network (green) to disentangle the feature map of a certain conv-layer in a pre-trained performer network (orange). The explainer network disentangles input features into object-part feature maps to explain knowledge representations in the performer, \emph{i.e.} making each filter represent a specific object part. The explainer network can also invert the disentangled object-part features to reconstruct features of the performer without much loss of information. We compare ordinary feature maps in the performer and the disentangled feature maps in the explainer on the right. The orange and green lines indicate information-pass routes for the inference process and the explanation process, respectively.}
\label{fig:top}
\end{figure}

\textbf{Tasks:} In order to solve the above dilemma, given a pre-trained CNN, we propose to learn an additional neural network, namely an \textit{explainer} network, to explain features inside the CNN. Accordingly, we call the pre-trained CNN a \textit{performer} network. As shown in Fig.~\ref{fig:top}, the explainer network explains feature maps of a certain conv-layer of the performer network. We attach the explainer onto the performer like an appendix without affecting the original discrimination power of the performer.

The explainer performs like an auto-encoder. The encoder in the explainer contains hundreds of interpretable filters, each of which disentangles an object part from input feature maps. The decoder inverts the disentangled object-part features to reconstruct features of upper layers of the performer.

As shown in Fig.~\ref{fig:top}, the feature map of each filter in the performer usually represents a chaotic mixture of object parts and textures, whereas as a paraphrase of performer features, the disentangled object-part features in the explainer provide an insightful understanding of the performer. People can obtain answers to the following two questions from the disentangled features.
\begin{enumerate}[itemindent=0em,listparindent=0em,leftmargin=1em]
\item What part patterns are encoded in the performer?
\item Which patterns are activated and used for each specific prediction?
\end{enumerate}

\textbf{Learning:} We learn the explainer by distilling feature representations from the performer to the explainer without any additional supervision. No annotations of sample labels, parts, or textures are used to guide feature disentanglement during the learning process. We add a loss to each interpretable filter in the explainer (see Fig.~\ref{fig:block}). The filter loss encourages an interpretable filter 1) to represent objects of a single category, and 2) to be triggered by a single region (part) of the object, rather than repetitively appear on different regions of an object. We assume that repetitive shapes on various regions are more likely to describe low-level textures (\emph{e.g.} colors and edges), than high-level parts\footnote[1]{\emph{E.g.}, we consider the left and right eyes as two different parts, because they have different contexts.}. Thus, the filter loss pushes each interpretable filter towards the representation of an object part.

Meanwhile, the disentangled object-part features are also required to reconstruct features of upper layers of the performer. Successful feature reconstructions can avoid significant information loss during the disentanglement of features.

\textbf{Potential values of the explainer:} The high interpretability of a middle conv-layer is of great value when the performer network needs to earn trust from people. As discussed in \cite{CNNBias}, due to potential dataset bias and representation bias, a high accuracy on testing images cannot fully ensure that the performer encodes correct features. People usually need clear explanations for middle-layer features of a performer, in order to semantically or visually clarify the logic of each prediction made by the performer. For example, in Fig.~\ref{fig:heatmap_analysis}, we use the disentangled object-part features to quantitatively identify which parts are learned and used for the prediction. Given an input image, some previous studies~\cite{visualCNN_grad,visualCNN_grad_2,trust} estimated the top contributing image regions in each prediction. In comparisons, our explainer ensures each interpretable filter to exclusively describe a single object part of a category, which provides more fine-grained knowledge structures inside the performer at the part level.

\textbf{Contributions:} We can summarize the contributions of this study as follows. 1) We tackle a new problem, \emph{i.e.} learning an explainer network to mine and clarify potential feature components that are encoded in middle layers of a pre-trained performer network. The explainer disentangles chaotic feature maps of the performer into human-interpretable object parts. Compared to directly letting the performer encode disentangled/interpretable features, learning an additional explainer does not affect the discrimination power of the performer, thereby ensuring broader applicability. 2) We propose a simple yet effective method to learn the explainer without any annotations of object parts or textures for supervision. 3) Theoretically, besides the proposed explainer, our explainer-performer network structure also supports knowledge distillation into new explainers with novel interpretability losses. Meanwhile, the explainer can be broadly applied to different CNN performers. 4) Experiments show that our approach has considerably improved the feature interpretability.

\section{Related work}

\textbf{Network interpretability:} Recent studies of the interpretability of visual network are reviewed in \cite{InterpretabilitySurvey}. Instead of analyzing network features from a global view~\cite{InformationBottleneck,InformationBottleneck2,CNNSpaceVisualization}, Bau~\emph{et al.}~\cite{Interpretability} defined six kinds of semantics for middle-layer feature maps of a CNN, \emph{i.e.} \textit{objects}, \textit{parts}, \textit{scenes}, \textit{textures}, \textit{materials}, and \textit{colors}. We can roughly consider the first two semantics as object-part patterns with specific shapes, and summarize the last four semantics as texture patterns. In this study, we use the explainer to disentangle object-part patterns from feature maps.

Many studies for network interpretability mainly visualized image appearance corresponding to a neural unit inside a CNN~\cite{CNNVisualization_1,CNNVisualization_2,CNNVisualization_3,FeaVisual,CNNVisualization_6,CNNVisualization_7} or extracted image regions that were responsible for network output~\cite{trust,trustRegion,CNNInfluence,visualCNN_grad,visualCNN_grad_2,ExplainingArea}. Other studies retrieved mid-level representations with specific meanings from CNNs for various applications~\cite{explainableFeature,explainableFeature2,explainableFeature3,explainableFeature4}. For example, Zhou~\emph{et al.}~\cite{CNNSemanticDeep,CNNSemanticDeep2} selected neural units to describe ``scenes''. Simon~\emph{et al.} discovered objects from feature maps of unlabeled images~\cite{ObjectDiscoveryCNN_2}. Zhang~\emph{et al.}~\cite{CNNAoG,DeepQA} extracted certain neural units to describe an object part in a weakly-supervised manner. Zhang~\emph{et al.}~\cite{explanatoryGraph} also disentangled feature representations in middle layers of a CNN into a graphical model of object parts in an unsupervised manner. In fact, each filter in a middle conv-layer usually encodes a mixture of parts and textures, and these studies consider the most notable part/texture component as the semantic meaning of a filter. In contrast, our research uses a filter loss to purify the semantic meaning of each filter (Fig.~\ref{fig:top} visualizes the difference between the two types of filters).

A new trend related to network interpretability is to learn networks with disentangled, interpretable representations~\cite{LogicRuleNetwork,CNNCompositionality,Parsimonious}. Many studies learn interpretable representations in a weakly-supervised or unsupervised manner. For example, capsule nets~\cite{capsule} and interpretable RCNN~\cite{InterRCNN} learned interpretable middle-layer features. InfoGAN~\cite{infoGAN} and $\beta$-VAE~\cite{betaVAE} learned meaningful input codes of generative networks. The study of interpretable CNNs~\cite{interpretableCNN} developed a loss to push each middle-layer filter towards the representation of a specific object part during the learning process without given part annotations, which is the closest to our research. However, as mentioned in \cite{Interpretability}, an interpretable model cannot always ensure a high discrimination power, which limits the applicability of above interpretable models. Therefore, instead of directly boosting the interpretability of the performer network, we propose to learn an explainer network in an unsupervised fashion.

\textbf{Meta-learning:} Our study is also related to meta-learning~\cite{meta1,meta2,meta3,meta4}. Meta-learning uses an additional model to guide the learning of the target model. In contrast, our research uses an additional explainer network to interpret middle-layer features of the target performer network.

\section{Algorithm}

\subsection{Explainer structure}

\begin{figure}[t]
\centering
\includegraphics[width=0.99\linewidth]{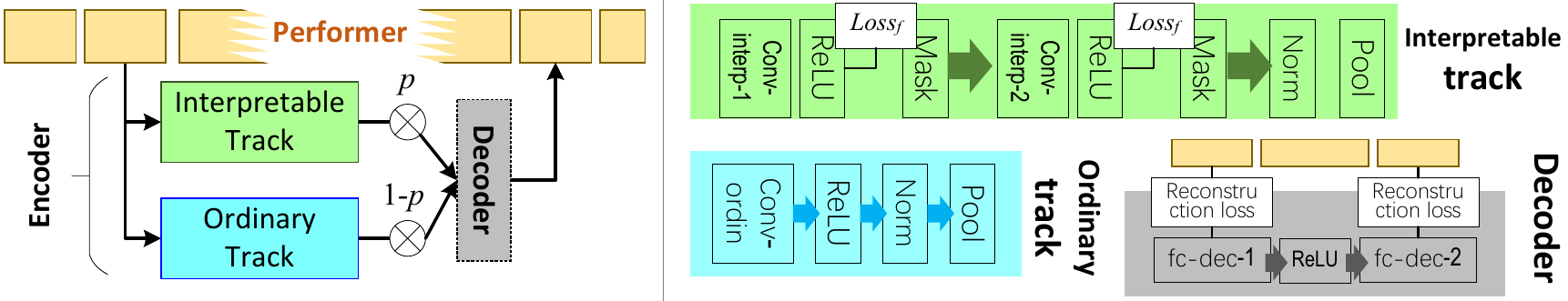}
\caption{Overall structure of the explainer network (left). Detailed structures within the interpretable track, the ordinary track, and the decoder are shown on the right. Theoretically, people can edit the number of conv-layers and FC layers within the encoder and the decoder for their own applications.}
\label{fig:block}
\end{figure}

As shown in Fig.~\ref{fig:block}, the explainer network has two modules, \emph{i.e.} an encoder and a decoder, which transform performer features into interpretable object-part features and invert object-part features back to features of the performer, respectively. We applied the encoder and decoder with following structures to all types of performers in all experiments. However, people can edit the number of conv-layers and fully-connected (FC) layers in their specific applications.

\textbf{Encoder:} The encoder contains two tracks, namely an \textit{interpretable track} and an \textit{ordinary track}. The interpretable track disentangles performer features into object parts. This track has two interpretable conv-layers (namely \textit{conv-interp-1,conv-interp-2}), each followed by a ReLU layer and a mask layer. The interpretable layer contains interpretable filters, and each interpretable filter can be only triggered by a specific object part. Since exclusively using object-part features is not enough to reconstruct performer features, we also design an ordinary track as a supplement to the interpretable track, in order to represent textural features that cannot be modeled by the interpretable track. The ordinary track contains a conv-layer (namely \textit{conv-ordin}), a ReLU layer, and a pooling layer.

We sum up output features of the interpretable track {$x_{\textrm{interp}}$} and those of the ordinary track {$x_{\textrm{ordin}}$} as the final output of the encoder, \emph{i.e.} {$x_{\textrm{enc}}=p\cdot x_{\textrm{interp}}+(1-p)\cdot x_{\textrm{ordin}}$}, where a scalar weight $p$ measures the quantitative contribution from the interpretable track. $p$ is parameterized as a softmax probability {$p=sigmoid(w_{p})$}, $w_{p}\in{\boldsymbol\theta}$, where ${\boldsymbol\theta}$ is the set of parameters to be learned. Our method encourages a large $p$ so that most information in {$x_{\textrm{enc}}$} comes from the interpretable track.

\textit{Norm-layer:} We normalize {$x_{\textrm{interp}}$} and {$x_{\textrm{ordin}}$} to make the probability $p$ accurately represent the ratio of the contribution from the interpretable track, \emph{i.e.} making each channel of these feature maps produces the same magnitude of activation values. Thus, we add two norm-layers to the interpretable track and the ordinary track (see Fig.~\ref{fig:block}). For each input feature map {$x\in\mathbb{R}^{L\times L\times D}$}, the normalization operation is given as {$\hat{x}^{(ijk)}=x^{(ijk)}/\alpha_{k}$}, where {$\alpha_{k}\in{\boldsymbol\alpha}\subset{\boldsymbol\theta}$} denotes the average activation magnitude of the $k$-th channel {$\alpha_{k}=\mathbb{E}_{x}[\sum_{ij}\max(x^{(ijk)},0)]$} through feature maps of all images, where {$x^{(ijk)}$} denotes an element in the tensor $x$. We can update ${\boldsymbol\alpha}$ during the learning process, which is similar to the learning for batch normalization.

\textit{Mask layer:} We add mask layers after each of the two interpretable conv-layers to further remove noisy activations that are unrelated to the target object part. Let {$x_{f}\in\mathbb{R}^{L\times L}$} denote the feature map of an interpretable filter $f$ after the ReLU operation. The mask layer localizes the potential target object part on $x_{f}$ as the neural unit with the strongest activation {$\hat{\mu}={\arg\!\max}_{\mu=[i,j]}x_{f}^{(ij)}$}, where $\mu=[i,j]$ denotes the coordinate of a neural unit in $x_{f}$ ($1\leq i,j\leq L$), and $x_{f}^{(ij)}$ indicates the activation value of this unit. Based on the part-localization result $\hat{\mu}$, the mask layer assigns a mask $mask_{f}$ to $x_{f}$ to remove noises. \emph{I.e.} {$x_{f}^{\textrm{masked}}=x_{f}\circ mask_{f}$} is the output feature map, where $\circ$ denotes the Hadamard (element-wise) product. The mask \emph{w.r.t.} $\hat{\mu}$ is given as {$mask_{f}=\max(T_{\hat{\mu}},0)$}, where {$T_{\hat{\mu}}$} is a pre-define template that will be introduced later. When {$mask_{f}$} is determined, we treat {$mask_{f}$} as a constant to enable gradient back-propagation.

\textbf{Decoder:} The decoder inverts {$x_{\textrm{enc}}$} to {$x_{\textrm{dec}}$}, which reconstructs performer features. The decoder has two FC layers, which followed by two ReLU layers. We use the two FC layers, namely \textit{fc-dec-1} and \textit{fc-dec-2}, to reconstruct feature maps of two corresponding FC layers in the performer. The reconstruction loss will be introduced later. The better reconstruction of the FC features indicates that the explainer loses less information during the computation of {$x_{\textrm{enc}}$}.

\subsection{Learning}

When we distill knowledge representations from the performer to the explainer, we consider the following three terms: 1) the quality of network distillation, \emph{i.e.} the explainer needs to well reconstruct feature maps of upper layers in the performer, thereby minimizing the information loss; 2) the interpretability of feature maps of the interpretable track, \emph{i.e.} each filter in \textit{conv-interp-2} should exclusively represent a certain object part; 3) the relative contribution of the interpretable track \emph{w.r.t.} the ordinary track, \emph{i.e.} we hope the interpretable track to make much more contribution to the final CNN prediction than the ordinary track. Therefore, we design the following loss for each input image to learn the explainer.
\begin{equation}
\min_{\boldsymbol\theta}Loss,\qquad Loss={\sum}_{l\in{\bf L}}\lambda_{(l)}\Vert x_{(l)}-x_{(l)}^{*}\Vert^2-\eta\log p+{\sum}_{f}\lambda_{f}\cdot Loss_{f}(x_{f})
\end{equation}
where ${\boldsymbol\theta}$ denotes the set of parameters to be learned, including filter weights of conv-layers and FC layers in the explainer, $w_{p}$ for $p$, and ${\boldsymbol\alpha}$ for norm-layers. $\lambda_{(l)},\lambda_{f}$ and $\eta$ are scalar weights.

\textbf{The first term} {$\Vert x_{(l)}-x_{(l)}^{*}\Vert^2$} is the reconstruction loss, where $x_{(l)}$ denotes the feature of the FC layer $l$ in the decoder, {${\bf L}=\{fc-dec-1,fc-dec-2\}$}. {$x_{(l)}^{*}$} indicates the corresponding feature in the performer. \textbf{The second term} $-\log p$ encourages the interpretable track to make more contribution to the CNN prediction. \textbf{The third term} {$Loss_{f}(x_{f})$} is the loss of filter interpretability. Without annotations of object parts, the filter loss forces $x_{f}$ to be exclusively triggered by a specific object part of a certain category. As shown in Fig.~\ref{fig:block}, we add a filter loss to each interpretable filter $f$ in the two conv-layers (\textit{conv-interp-1} and \textit{conv-interp-2}). {$x_{f}\in\mathbb{R}^{L\times L}$} denotes the feature map of the interpretable filter after the ReLU operation. Note that $w_{p}$ for $p$ is updated based on both the $-\log p$ loss and other losses. The $-\log p$ loss encourages a high value of $p$, while the reconstruction loss usually requires a moderate value of $p$ to balance neural activations from the interpretable and ordinary tracks to ensure a good reconstruction.

%\subsubsection{Details of the learning process}
%During the forward propagation, the interpretable filter $f$ and the ReLU operation first produces a feature map $x_{f}$. Then, the mask layer localizes the potential target object part on $x_{f}$ and assigns a mask to remove unrelated activations. During the back propagation, the interpretable filter $f$ receives gradients from both the filter loss and the reconstruction loss to update its weights. Whereas, ordinary filters in the ordinary track and the decoder only learn from gradients of the reconstruction loss.

\begin{figure}
\centering
\includegraphics[width=0.95\linewidth]{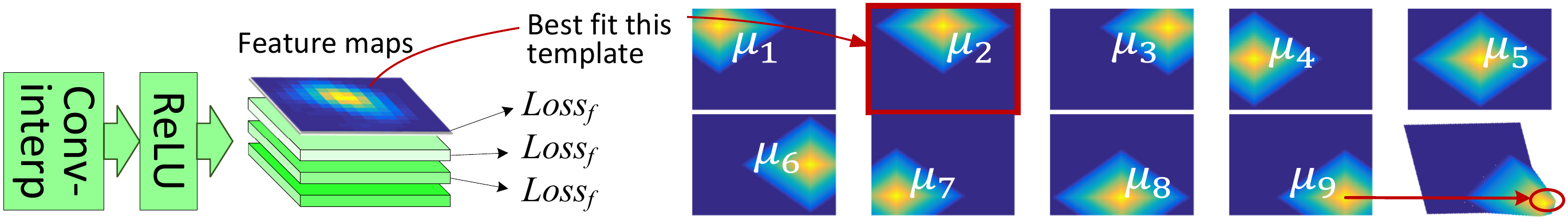}
\caption{Templates of {$\{T_{\mu_{i}}\}$} for the toy case of {$L=3$}. Each template {$T_{\mu_{i}}$} represents the ideal activation shape when the target part is localized at the $i$-th unit of $x_{f}$. In real applications, we usually use $6^2$--$14^2$ templates since {$6\leq L\leq 14$}.}
\label{fig:template}
\end{figure}

\subsubsection{Filter loss}

As a prior measurement of the fitness between a filter and an object part, the filter loss was formulated in \cite{interpretableCNN} as the minus mutual information between feature maps and a set of pre-defined templates.
\begin{equation}
{\bf Loss}_{f}={\sum}_{x_{f}\in{\bf X}}Loss_{f}(x_{f})=-MI({\bf X};{\bf T})=-{\sum}_{T\in{\bf T}}p(T){\sum}_{x_{f}\in{\bf X}}p(x_{f}|T)\log\frac{p(x_{f}|T)}{p(x_{f})}
\label{eqn:loss}
\end{equation}
where {$MI(\cdot)$} indicates the mutual information. {${\bf X}$} denotes a set of feature maps of the filter $f$, which are extracted from different input images. {${\bf T}=\{T_{\mu_1},T_{\mu_2},\ldots,T_{\mu_{L^2}},T^{-}\}$} is referred to as a set of pre-defined templates. {$p(T)$} is a constant prior probability of a template {$T$}, and {$p(x_{f}|T)$} measures the fitness between $x_{f}$ and {$T$}. Thus, we get {$Loss_{f}(x_{f})=-\sum_{T\in{\bf T}}p(x_{f},T)\log\frac{p(x_{f}|T)}{p(x_{f})}$}.

The first {$L^2$} templates, \emph{i.e.} {$T_{\mu_1},T_{\mu_2},\ldots,T_{\mu_{L^2}}$}. {$T_{\mu_{i}}\in\mathbb{R}^{L\times L}$} represents the ideal distributions of neural activations in the feature map $x_{f}$ when the target part mainly triggers the unit $\mu_{i}$ in $x_{f}$ ({$i\in\{1,2,\ldots,L^2\}$}). Fig.~\ref{fig:template} illustrates the {$L^2$} templates for a toy case of $L=3$. Besides the {$L^2$} positive templates, we use a negative template {$T^{-}$} ({$T^{-,(ij)}=-\tau$}) for the case that the input image does not contain the target object so that the feature map should remain inactivated. $\tau$ is a positive constant. \textbf{1)} When the input image contains the target object, we expect $x_{f}$ to be activated by a single part at a single location in the feature map. Thus, $x_{f}$ should match only one of the {$L^2$} positive templates. \textbf{2)} When the input image does not contain the target object, we expect $x_{f}$ not to be triggered, thereby matching the negative template {$T^{-}$}. In this way, for each input image, its feature map $x_{f}$ needs to match exactly one of the {$(L^2+1)$} templates. Thus, high mutual information in Equation~(\ref{eqn:loss}) indicates a high probability that the filter $f$ represents the same object part through different images.

More specifically, {$p(x_{f}|T)$} is given as {$p(x_{f}|T)=\frac{1}{Z_{T}}\exp[tr(x_{f}\cdot T)]$}, where {$Z_{T}=\sum_{x_{f}\in{\bf X}}\exp[tr(x_{f}\cdot T)]$}. {$tr(\cdot)$} indicates the trace of a matrix, \emph{i.e.} {$tr(x_{f}\cdot T)=\sum_{ij}x_{f}^{(ij)}T^{(ji)}$}. {$p(x_{f})=\sum_{T}p(T)p(x_{f}|T)$}.

\section{Experiments}

In experiments, we learned explainers for performer networks with three types of structures to demonstrate the broad applicability of our method. Performer networks were pre-trained using object images in two different benchmark datasets for object classification. We visualized feature maps of interpretable filters in the explainer to illustrate semantic meanings of these filters. Experiments showed that interpretable filters in the explainer generated more semantically meaningful feature maps than conv-layers in the pre-trained performer.

\textbf{Benchmark datasets:} Because the evaluation of filter interpretability required ground-truth annotations of object landmarks\footnote[2]{To avoid ambiguity, a landmark is referred to as the central position of a semantic part with an explicit name (\emph{e.g.} a head, a tail). In contrast, the part corresponding to an interpretable filter does not have an explicit name. We followed experiment settings in \cite{interpretableCNN}, which selected the \textit{head}, \textit{neck}, and \textit{torso} of each category in the Pascal-Part dataset~\cite{SemanticPart} as the landmarks and used the \textit{head}, \textit{back}, \textit{tail} of birds in the CUB200-2011 dataset~\cite{CUB200} as landmarks. It was because these landmarks appeared on testing images most frequently.} (parts), we used two benchmark datasets with part annotations for training and testing, \emph{i.e.} the CUB200-2011 dataset~\cite{CUB200} and the Pascal-Part dataset~\cite{SemanticPart}. Note that previous studies~\cite{SemanticPart,CNNAoG,interpretableCNN} usually selected animal categories to test part localization, because animals usually contain non-rigid parts, which present great challenges for part localization. Therefore, we followed the experimental design in \cite{CNNAoG,interpretableCNN} that selected the seven animal categories in the two datasets for evaluation. Both the datasets provide object bounding boxes. The CUB200-2011 dataset~\cite{CUB200} contains 11.8K bird images of 200 species with center positions of fifteen bird landmarks. Here, we considered all 200 bird species in the CUB200-2011 dataset as a single category. The Pascal-Part dataset~\cite{SemanticPart} provides ground-truth segmentations of a total of 107 object parts for six animal categories.

\textbf{Four types of CNNs as performers:} We applied our method to four types of performers, including the AlexNet~\cite{CNNImageNet}, the VGG-M network~\cite{VGG}, the VGG-S network~\cite{VGG}, the VGG-16 network~\cite{VGG}. For residual networks~\cite{ResNet} and dense networks~\cite{denseNet}, skip connections in these networks usually make a single feature map contain rich features from different conv-layers, which increases the difficulty of understanding its feature maps. Thus, to simplify the story, we did not test the performance on residual networks and dense networks.

\textbf{Two experiments:} We followed experimental settings in \cite{interpretableCNN} to conduct two experiments, \emph{i.e.} an experiment of single-category classification and an experiment of multi-category classification. For single-category classification, we learned four performers with structures of the AlexNet~\cite{CNNImageNet}, VGG-M~\cite{VGG}, VGG-S~\cite{VGG}, and VGG-16~\cite{VGG} for each of the seven animal categories in the two benchmark datasets. Thus, we learned 28 performers, and each performer was learned to classify objects of a certain category from other objects. We cropped objects of the target category based on their bounding boxes as positive samples. Images of other categories were regarded as negative samples. For multi-category classification, we learned the VGG-M~\cite{VGG}, VGG-S~\cite{VGG}, and VGG-16~\cite{VGG} to classify the six animal categories in the Pascal-Part dataset~\cite{SemanticPart}.

\textbf{Experimental details:} We considered the \textit{relu4} layer of the AlexNet/VGG-M/VGG-S (the 12th/12th/11th layer of the AlexNet/VGG-M/VGG-S) and the \textit{relu5-2} layer of the VGG-16 (the 28th layer) as target layers. We sent feature maps of the target layer in each performer network to an explainer network. We learned the explainer network to disentangle these feature maps for testing. The output of the explainer reconstructed the feature of the \textit{fc7} layer of the performer (the 19th/19th/18th/35th layer of the AlexNet/VGG-M/VGG-S/VGG-16) and was fed back to the performer. Thus, a reconstruction loss matched features between the \textit{fc-dec-2} layer of the explainer and the \textit{fc7} layer of the performer. Another reconstruction loss connected the \textit{fc-dec-1} layer of the explainer and the previous \textit{fc6} layer of the performer. Each conv-layer in the explainer had {$D$} filters with a {$3\times3\times D$} kernel and a biased term, where {$D$} is the channel number of its input feature map. We used zero padding to ensure the output feature map had the same size of the input feature map. We initialized the \textit{conv-interp-2} and \textit{conv-ordin} layers with random weights. The \textit{conv-interp-1} layer in the explainer was initialized using weights of the \textit{conv5}/\textit{conv5}/\textit{conv5}/\textit{conv5-3} layer of the AlexNet/VGG-M/VGG-S/VGG-16. The \textit{fc-dec-1} and \textit{fc-dec-2} layers in the explainer copied filter weights from the \textit{fc6} and \textit{fc7} layers of the performer, respectively. Pooling layers in the explainer were also parameterized considering the last pooling layer in the performer.

Theoretically, we can use different activation shapes for {$T_{\mu_{i}}$} to construct the filter loss. For simplicity, we used part templates defined in \cite{interpretableCNN} in our experiments. We set $\eta=1.0\times 10^6$ for the AlexNet, VGG-M, and VGG-S and set $\eta=1.5\times 10^5$ for the VGG-16, since the VGG-16 has more conv-layers than the other networks. We set {$\lambda_{(l)}=5\times 10^4/\mathbb{E}_{x_{(l)}^{*}}[\Vert \max(x_{(l)}^{*},0)\Vert]$} in all experiments, where the expectation was averaged over features of all images. We followed \cite{interpretableCNN} to set templates {${\bf T}$} with the parameter $\tau$ and {$p(T)=\frac{1}{L^2+1}$}. Directly minimizing {$Loss_{f}(x_{f})$} is time-consuming, and \cite{interpretableCNN} formulated approximate gradients of {$\lambda_{f}Loss_{f}(x_{f})$} to speed up the computation. Please see the appendix for more details.

%% parameter settings

\textbf{Evaluation metric:} We compared the object-part interpretability between feature maps of the explainer and those of the performer. According to above network structures, we can parallel the explainer to the top conv-layer of the performer, because they both receive features from the \textit{relu4}/\textit{relu5-2} layer of the performer and output features to the upper layers of the performer. Crucially, as discussed in \cite{Interpretability}, low conv-layers in a CNN usually represent colors and textures, while high conv-layers mainly represent object parts; the top conv-layer of the CNN is most likely to model object parts among all conv-layers. Therefore, to enable a fair comparison, we compared feature maps of the \textit{conv-interp-2} layer of the explainer with feature maps of the top conv-layer of the performer.

We used the location instability as the evaluation metric, which has been widely used to measure the fitness between a filter $f$ and the representation of a specific object part~\cite{interpretableCNN}. Given a feature map $x_{f}$, we localized the part at the unit $\hat{\mu}$ with the highest activation. We used \cite{CNNSemanticDeep} to project the part coordinate $\hat{\mu}$ on the feature map onto the image plane and obtained ${\bf p}_{\hat{\mu}}$. We assumed that if the filter $f$ consistently represented the same object part of a certain category through different images, then distances between the inferred part location ${\bf p}_{\hat{\mu}}$ and some object landmarks\textcolor{red}{\footnotemark[2]} of the category should not change a lot among different objects. For example, if $f$ always represented the head part on different objects, then the distance between the localized part ${\bf p}_{\hat{\mu}}$ (\emph{i.e.} the dog head) and the ground-truth landmark of the shoulder should keep stable, although the head location ${\bf p}_{\hat{\mu}}$ may change in different images. Thus, for single-category classification, we computed the deviation of the distance between ${\bf p}_{\hat{\mu}}$ and a specific landmark through objects of the category, and we used the average deviation \emph{w.r.t.} various landmarks to evaluate the location instability of $f$. The location instability was reported as the average deviation, when we computed deviations using all pairs of filters and landmarks of the category. For multi-category classification, we first determined the target category of each filter $f$ and then computed the location instability based on objects of the target category. We assigned each interpretable filter in the explainer to the category whose images can activate the filter most. Please see \cite{interpretableCNN} for computational details of this evaluation metric.

%We assigned each ordinary filter in the performer to the category whose landmarks can achieve the lowest location deviation to simplify the computation.

\begin{table}
\caption{Location instability of feature maps in performers and explainers for the evaluation of filter interpretability. Please see the appendix for comparisons with more baselines.}
\label{tab:instability}
\noindent$\!\!\!$
\resizebox{9.7cm}{!}{\begin{tabular}{l|ccccccc|c}
\multicolumn{8}{c}{\large Results based on the Pascal-Part dataset~\cite{SemanticPart}}\\
\hline
&\multicolumn{7}{|c|}{\large Single-category} & Multi-category\\
\hline
& bird & cat & cow & dog & horse & sheep & Avg. & Avg.\\
\hline
AlexNet &0.153
&0.131
&0.141
&0.128
&0.145
&0.140
&\textcolor{blue}{0.140}
& --\\
Explainer &{\bf0.104}
&{\bf0.089}
&{\bf0.101}
&{\bf0.083}
&{\bf0.098}
&{\bf0.103}
&\textcolor{blue}{\bf0.096}
& --\\
\hline
VGG-M &0.152
&0.132
&0.143
&0.130
&0.145
&0.141
&\textcolor{blue}{0.141}
&0.135\\
Explainer &{\bf0.106}
&{\bf0.088}
&{\bf0.101}
&{\bf0.088}
&{\bf0.097}
&{\bf0.101}
&\textcolor{blue}{\bf0.097}
&{\bf0.097}\\
\hline
VGG-S &0.152
&0.131
&0.141
&0.128
&0.144
&0.141
&\textcolor{blue}{0.139}
&0.138\\
Explainer &{\bf0.110}
&{\bf0.085}
&{\bf0.098}
&{\bf0.085}
&{\bf0.091}
&{\bf0.096}
&\textcolor{blue}{\bf0.094}
&{\bf0.107}\\
\hline
VGG-16 &0.145
&0.133
&0.146
&0.127
&0.143
&0.143
&\textcolor{blue}{0.139}
& 0.128\\
Explainer &{\bf0.095}
&{\bf0.089}
&{\bf0.097}
&{\bf0.085}
&{\bf0.087}
&{\bf0.089}
&\textcolor{blue}{\bf0.090}
&{\bf0.109}\\
\hline
\end{tabular}}
\quad
\resizebox{3.6cm}{!}{\begin{tabular}{cc}
\multicolumn{2}{c}{CUB200-2011 dataset~\cite{CUB200}}\\
\\
\hline
AlexNet & 0.1502\\
Explainer & {\bf0.0906}\\
\hline
VGG-M & 0.1476\\
Explainer & {\bf0.0815}\\
\hline
VGG-S & 0.1481\\
Explainer & {\bf0.0704}\\
\hline
VGG-16 & 0.1373\\
Explainer& {\bf0.0490}\\
\hline
\end{tabular}}
\end{table}

\begin{table}
\begin{floatrow}
\noindent$\!\!\!$
\resizebox{5.9cm}{!}{\capbtabbox{\begin{tabular}{l|c|c|c}
\hline
& \multicolumn{2}{c|}{Pascal-Part~\cite{SemanticPart}} & CUB200\\
\!&\!\!\! Single \!\!\!&\!\!\! Multi \!\!\!&\!\!\! -2011~\cite{CUB200}\\
\hline
\!\!{\footnotesize AlexNet}
&--
&0.7137
&0.5810\\
\!\!{\footnotesize VGG-M}
&0.9012
&0.8066
&0.8611\\
\!\!{\footnotesize VGG-S}
&0.9270
&0.8996
&0.9533\\
\!\!{\footnotesize VGG-16}
&0.8593
&0.8718
&0.9579\\
\hline
\end{tabular}
}{\caption{\large Average $p$ values of explainers in different experiments.}
\label{tab:p}}}
\resizebox{8cm}{!}{\capbtabbox{\vspace{25pt}\begin{tabular}{l|c|cc|cc}
\hline
\!\!\!&\!\!\! {\footnotesize Performer} \!\!\!&\!\!\! {\footnotesize Explainer} \!\!\!&\!\!\! $\Delta$ Error \!\!\!&\!\!\! {\footnotesize Performer+cls} \!\!\!&\!\!\! $\Delta$ Error \!\!\!\\
\hline
\!\!\!VGG-M \!\!\!&\!\!\! 6.12\% \!\!\!&\!\!\! 6.62\% \!\!\!&\!\!\! 0.5\% \!\!\!&\!\!\! 5.22\% \!\!\!&\!\!\! -0.9\% \!\!\!\\
\!\!\!VGG-S \!\!\!&\!\!\! 5.95\% \!\!\!&\!\!\! 6.97\% \!\!\!&\!\!\! 1.02\% \!\!\!&\!\!\! 5.43\% \!\!\!&\!\!\! -0.52\% \!\!\!\\
\!\!\!VGG-16 \!\!\!&\!\!\! 2.03\% \!\!\!&\!\!\! 2.17\% \!\!\!&\!\!\! 0.14\% \!\!\!&\!\!\! 2.49\% \!\!\!&\!\!\! 0.46\% \!\!\!\\
\hline
\end{tabular}
}{\caption{Multi-category classification errors using features of performers and explainers based on the Pascal-Part dataset~\cite{SemanticPart}. We evaluated loss of feature information in explainers based on the classification-error gap between performers and explainers. Please see the appendix for more results.\vspace{-30pt}}
\label{tab:classification}}}
\end{floatrow}
\end{table}

\begin{figure}[t]
\centering
\includegraphics[width=0.99\linewidth]{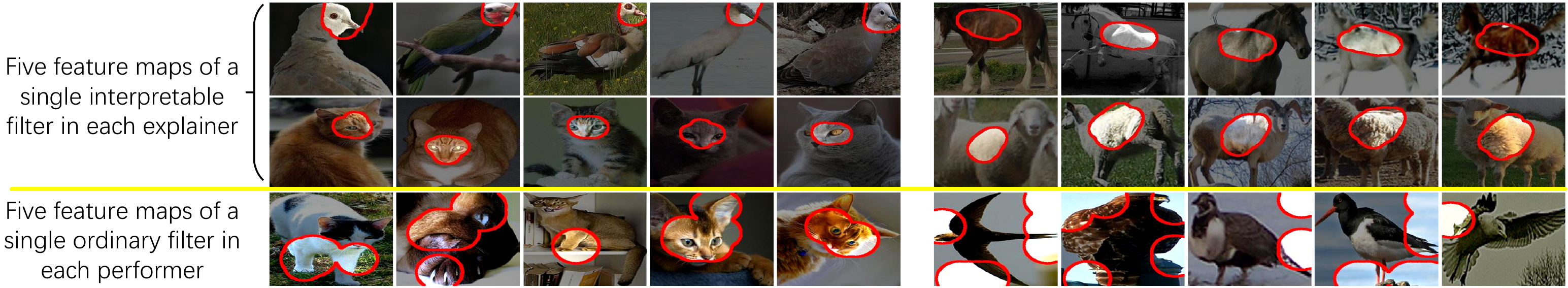}
\caption{Visualization of interpretable filters in the explainer and ordinary filters in the performer. As discussed in \cite{Interpretability}, the top conv-layer of a CNN is the more likely to represent object parts than low conv-layers. We visualized and compared filters in the top conv-layer of the performer and interpretable filters in the \textit{conv-interp-2} layer of the explainer. We used \cite{CNNSemanticDeep} to estimate the RF\textcolor{red}{{\protect\footnotemark[3]}} of neural activations to illustrate a filter's semantics. Interpretable filters are much more semantically meaningful than ordinary filters. Please see the appendix for more results.}
\label{fig:visual}
\end{figure}

\subsection{Experimental results and analysis}

Table~\ref{tab:instability} compares the interpretability between feature maps in the performer and feature maps in the explainer. Feature maps in our explainers were much more interpretable than feature maps in performers in all comparisons. The explainer exhibited only 35.7\%--68.8\% of the location instability of the performer, which means that interpretable filters in the explainer more consistently described the same object part through different images than filters in the performer. The $p$ value of an explainer indicates the quantitative ratio of the contribution from interpretable features. Table~\ref{tab:p} lists $p$ values of explainers that were learned for different performers. When we used an explainer to interpret feature maps of a VGG network, about 80\%--96\% activation scores came from interpretable features. To evaluate feature reconstructions of an explainer, we fed the reconstructed features back to the performer for classification. As shown in Table~\ref{tab:classification}, we compared the classification accuracy of explainer's reconstructed features with the accuracy based on original performer features. Performers outperformed explainers in object classification. We used the explainer's increase of classification errors \emph{w.r.t.} the performer (\emph{i.e.} ``$\Delta$ Error'' in Table~\ref{tab:classification}) to measure the information loss during feature transformation in the explainer. Furthermore, we added another baseline (\textit{Explainer+cls}), which used the classification loss to replace the reconstruction loss to learned explainers. \textit{Explainer+cls} learned from the classification loss, rather than distill knowledge from performers, so we did not consider feature maps in \textit{Explainer+cls} as the explanation of performer features. However, with the help of the classification loss, \textit{Explainer+cls} outperformed the performer in classification, which indicates potential boundaries of the discrimination power of explainers.

\begin{figure}[t]
\centering
\includegraphics[width=0.99\linewidth]{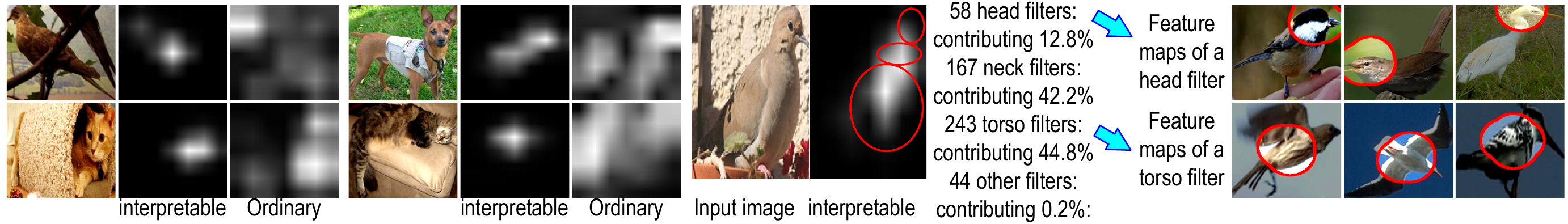}
\caption{Grad-CAM attention maps and quantitative analysis. We used \cite{visualCNN_grad_2} to compute grad-CAM attention maps of interpretable feature maps in the explainer and ordinary feature maps in the performer. Interpretable filters focused on a few distinct object parts, while ordinary filters separated its attention to both textures and parts. We can assign each interpretable filter with a semantic part. For example, we learned 58, 167, and 243 filters in the \textit{conv-interp-2} layer to represent the head, neck, and torso of the bird, respectively. We used \cite{explanatoryTree_arXiv} to compute quantitative contributions of different parts to the classification score on the right. Please see the appendix for more results.}
\label{fig:heatmap_analysis}
\end{figure}

\textbf{Visualization of filters:} We used the visualization method proposed by Zhou~\emph{et al.}~\cite{CNNSemanticDeep} to compute the receptive field (RF) of neural activations of an interpretable filter (after ReLU and mask operations), which was scaled up to the image resolution. As mentioned in \cite{CNNSemanticDeep}, the computed RF represented image regions that were responsible for neural activations of a filter, which was much smaller than the theoretical size of the RF. Fig.~\ref{fig:visual} used RFs\footnote[3]{When an ordinary filter in the performer does not have consistent contours, it is difficult for \cite{CNNSemanticDeep} to align different images to compute the average RF. Thus, for performers, we simply used a round RF for each activation. We overlapped all activated RFs in a feature map to compute the final RF.} to visualize interpretable filters in the \textit{conv-interp-2} layer of the explainer and ordinary filters in the top conv-layer of the performer. Fig.~\ref{fig:heatmap_analysis} compares grad-CAM attention maps~\cite{visualCNN_grad_2} of the \textit{conv-interp-2} layer in the explainer and those of the top conv-layer of the performer. Interpretable filters in an explainer mainly represented an object part, while feature maps of ordinary filters were usually activated by different image regions without clear semantic meanings.

\section{Conclusion and discussions}

In this paper, we have proposed a theoretical solution to a new task, \emph{i.e.} learning an explainer network to disentangle and explain feature maps of a pre-trained performer network. Learning an explainer besides the performer does not decrease the discrimination power of the performer, which ensures the broad applicability. We have developed a simple yet effective method to learn the explainer, which guarantees the high interpretability of feature maps without using annotations of object parts or textures for supervision. Theoretically, our explainer-performer structure supports knowledge distillation into new explainer networks with different losses. People can revise network structures inside the ordinary track, the interpretable track, and the decoder and apply novel interpretability losses to the interpretable track for each specific application.

We have applied our method to different types of performers, and experimental results show that our explainers can disentangle most information of input feature maps into object-part feature maps, which significantly boosts the feature interpretability. \emph{E.g.} for explainers for VGG networks, more than 80\% signals came from interpretable filters. The explainer can also invert object-part feature maps back to reconstruct feature maps of the performer without losing much information.

{\small
\bibliographystyle{ieee}
\bibliography{TheBib}

\begin{thebibliography}{10}\itemsep=-1pt

\bibitem{meta2}
M.~Andrychowicz, M.~Denil, S.~G. Colmenarejo, M.~W. Hoffman, D.~Pfau,
  T.~Schaul, B.~Shillingford, and N.~de~Freitas.
\newblock Learning to learn by gradient descent by gradient descent.
\newblock {\em In NIPS}, 2016.

\bibitem{Interpretability}
D.~Bau, B.~Zhou, A.~Khosla, A.~Oliva, and A.~Torralba.
\newblock Network dissection: Quantifying interpretability of deep visual
  representations.
\newblock {\em In CVPR}, 2017.

\bibitem{infoGAN}
X.~Chen, Y.~Duan, R.~Houthooft, J.~Schulman, I.~Sutskever, and P.~Abbeel.
\newblock Infogan: Interpretable representation learning by information
  maximizing generative adversarial nets.
\newblock {\em In NIPS}, 2016.

\bibitem{SemanticPart}
X.~Chen, R.~Mottaghi, X.~Liu, S.~Fidler, R.~Urtasun, and A.~Yuille.
\newblock Detect what you can: Detecting and representing objects using
  holistic models and body parts.
\newblock {\em In {CVPR}}, 2014.

\bibitem{meta1}
Y.~Chen, M.~W. Hoffman, S.~G. Colmenarejo, M.~Denil, T.~P. Lillicrap,
  M.~Botvinick, and N.~de~Freitas.
\newblock Learning to learn without gradient descent by gradient descent.
\newblock {\em In ICML}, 2017.

\bibitem{CNNVisualization_7}
Y.~Dong, H.~Su, J.~Zhu, and F.~Bao.
\newblock Towards interpretable deep neural networks by leveraging adversarial
  examples.
\newblock {\em In arXiv:1708.05493}, 2017.

\bibitem{FeaVisual}
A.~Dosovitskiy and T.~Brox.
\newblock Inverting visual representations with convolutional networks.
\newblock {\em In {CVPR}}, 2016.

\bibitem{trustRegion}
E.~R. Elenberg, A.~G. Dimakis, M.~Feldman, and A.~Karbasi.
\newblock Streaming weak submodularity: Interpreting neural networks on the
  fly.
\newblock {\em In NIPS}, 2017.

\bibitem{visualCNN_grad}
R.~C. Fong and A.~Vedaldi.
\newblock Interpretable explanations of black boxes by meaningful perturbation.
\newblock {\em In arXiv:1704.03296v1}, 2017.

\bibitem{ResNet}
K.~He, X.~Zhang, S.~Ren, and J.~Sun.
\newblock Deep residual learning for image recognition.
\newblock {\em In {CVPR}}, 2016.

\bibitem{betaVAE}
I.~Higgins, L.~Matthey, A.~Pal, C.~Burgess, X.~Glorot, M.~Botvinick,
  S.~Mohamed, and A.~Lerchner.
\newblock $\beta$-vae: learning basic visual concepts with a constrained
  variational framework.
\newblock {\em In ICLR}, 2017.

\bibitem{LogicRuleNetwork}
Z.~Hu, X.~Ma, Z.~Liu, E.~Hovy, and E.~P. Xing.
\newblock Harnessing deep neural networks with logic rules.
\newblock {\em In {arXiv:1603.06318v2}}, 2016.

\bibitem{denseNet}
G.~Huang, Z.~Liu, K.~Q. Weinberger, and L.~van~der Maaten.
\newblock Densely connected convolutional networks.
\newblock {\em In CVPR}, 2017.

\bibitem{CNNInfluence}
P.~Koh and P.~Liang.
\newblock Understanding black-box predictions via influence functions.
\newblock {\em In ICML}, 2017.

\bibitem{explainableFeature}
S.~Kolouri, C.~E. Martin, and H.~Hoffmann.
\newblock Explaining distributed neural activations via unsupervised learning.
\newblock {\em In CVPR Workshop on Explainable Computer Vision and Job
  Candidate Screening Competition}, 2017.

\bibitem{CNNImageNet}
A.~Krizhevsky, I.~Sutskever, and G.~E. Hinton.
\newblock Imagenet classification with deep convolutional neural networks.
\newblock {\em In {NIPS}}, 2012.

\bibitem{ExplainingArea}
D.~Kumar, A.~Wong, and G.~W. Taylor.
\newblock Explaining the unexplained: A class-enhanced attentive response
  (clear) approach to understanding deep neural networks.
\newblock {\em In CVPR Workshop on Explainable Computer Vision and Job
  Candidate Screening Competition}, 2017.

\bibitem{CNN}
Y.~LeCun, L.~Bottou, Y.~Bengio, and P.~Haffner.
\newblock Gradient-based learning applied to document recognition.
\newblock {\em In {Proceedings of the IEEE}}, 1998.

\bibitem{explainableFeature4}
B.~J. Lengerich, S.~Konam, E.~P. Xing, S.~Rosenthal, and M.~Veloso.
\newblock Visual explanations for convolutional neural networks via input
  resampling.
\newblock {\em In ICML Workshop on Visualization for Deep Learning}, 2017.

\bibitem{meta3}
K.~Li and J.~Malik.
\newblock Learning to optimize.
\newblock {\em In arXiv:1606.01885}, 2016.

\bibitem{Parsimonious}
R.~Liao, A.~Schwing, R.~Zemel, and R.~Urtasun.
\newblock Learning deep parsimonious representations.
\newblock {\em In NIPS}, 2016.

\bibitem{CNNVisualization_2}
A.~Mahendran and A.~Vedaldi.
\newblock Understanding deep image representations by inverting them.
\newblock {\em In CVPR}, 2015.

\bibitem{CNNSpaceVisualization}
P.~E. Rauber, S.~G. Fadel, A.~X.~F. {a}o, and A.~C. Telea.
\newblock Visualizing the hidden activity of artificial neural networks.
\newblock {\em In Transactions on PAMI}, 23(1):101--110, 2016.

\bibitem{trust}
M.~T. Ribeiro, S.~Singh, and C.~Guestrin.
\newblock ``why should i trust you?'' explaining the predictions of any
  classifier.
\newblock {\em In KDD}, 2016.

\bibitem{capsule}
S.~Sabour, N.~Frosst, and G.~E. Hinton.
\newblock Dynamic routing between capsules.
\newblock {\em In NIPS}, 2017.

\bibitem{InformationBottleneck2}
R.~Schwartz-Ziv and N.~Tishby.
\newblock Opening the black box of deep neural networks via information.
\newblock {\em In arXiv:1703.00810}, 2017.

\bibitem{visualCNN_grad_2}
R.~R. Selvaraju, M.~Cogswell, A.~Das, R.~Vedantam, D.~Parikh, and D.~Batra.
\newblock Grad-cam: Visual explanations from deep networks via gradient-based
  localization.
\newblock {\em In ICCV}, 2017.

\bibitem{ObjectDiscoveryCNN_2}
M.~Simon and E.~Rodner.
\newblock Neural activation constellations: Unsupervised part model discovery
  with convolutional networks.
\newblock {\em In {ICCV}}, 2015.

\bibitem{CNNVisualization_3}
K.~Simonyan, A.~Vedaldi, and A.~Zisserman.
\newblock Deep inside convolutional networks: visualising image classification
  models and saliency maps.
\newblock {\em In {arXiv:1312.6034}}, 2013.

\bibitem{VGG}
K.~Simonyan and A.~Zisserman.
\newblock Very deep convolutional networks for large-scale image recognition.
\newblock {\em In {ICLR}}, 2015.

\bibitem{CNNCompositionality}
A.~Stone, H.~Wang, Y.~Liu, D.~S. Phoenix, and D.~George.
\newblock Teaching compositionality to cnns.
\newblock {\em In CVPR}, 2017.

\bibitem{explainableFeature3}
C.~Ventura, D.~Masip, and A.~Lapedriza.
\newblock Interpreting cnn models for apparent personality trait regression.
\newblock {\em In CVPR Workshop on Explainable Computer Vision and Job
  Candidate Screening Competition}, 2017.

\bibitem{CUB200}
C.~Wah, S.~Branson, P.~Welinder, P.~Perona, and S.~Belongie.
\newblock The caltech-ucsd birds-200-2011 dataset.
\newblock Technical report, In {California Institute of Technology}, 2011.

\bibitem{meta4}
J.~Wang, Z.~Kurth-Nelson, D.~Tirumala, H.~Soyer, J.~Leibo, R.~Munos,
  C.~Blundell, D.~Kumaran, and M.~Botvinick.
\newblock Learning to reinforcement learn.
\newblock {\em In arXiv:1611.05763v3}, 2017.

\bibitem{explainableFeature2}
A.~S. Wicaksana and C.~C.~S. Liem.
\newblock Human-explainable features for job candidate screening prediction.
\newblock {\em In CVPR Workshop on Explainable Computer Vision and Job
  Candidate Screening Competition}, 2017.

\bibitem{InformationBottleneck}
N.~Wolchover.
\newblock New theory cracks open the black box of deep learning.
\newblock {\em In Quanta Magazine}, 2017.

\bibitem{InterRCNN}
T.~Wu, X.~Li, X.~Song, W.~Sun, L.~Dong, and B.~Li.
\newblock Interpretable r-cnn.
\newblock {\em In arXiv:1711.05226}, 2017.

\bibitem{CNNVisualization_6}
J.~Yosinski, J.~Clune, A.~Nguyen, T.~Fuchs, and H.~Lipson.
\newblock Understanding neural networks through deep visualization.
\newblock {\em In ICML Deep Learning Workshop}, 2015.

\bibitem{CNNVisualization_1}
M.~D. Zeiler and R.~Fergus.
\newblock Visualizing and understanding convolutional networks.
\newblock {\em In {ECCV}}, 2014.

\bibitem{explanatoryGraph}
Q.~Zhang, R.~Cao, F.~Shi, Y.~Wu, and S.-C. Zhu.
\newblock Interpreting cnn knowledge via an explanatory graph.
\newblock {\em In AAAI}, 2018.

\bibitem{CNNAoG}
Q.~Zhang, R.~Cao, Y.~N. Wu, and S.-C. Zhu.
\newblock Growing interpretable part graphs on convnets via multi-shot
  learning.
\newblock {\em In {AAAI}}, 2016.

\bibitem{DeepQA}
Q.~Zhang, R.~Cao, Y.~N. Wu, and S.-C. Zhu.
\newblock Mining object parts from cnns via active question-answering.
\newblock {\em In CVPR}, 2017.

\bibitem{CNNBias}
Q.~Zhang, W.~Wang, and S.-C. Zhu.
\newblock Examining cnn representations with respect to dataset bias.
\newblock {\em In AAAI}, 2018.

\bibitem{interpretableCNN}
Q.~Zhang, Y.~N. Wu, and S.-C. Zhu.
\newblock Interpretable convolutional neural networks.
\newblock {\em In CVPR}, 2018.

\bibitem{explanatoryTree_arXiv}
Q.~Zhang, Y.~Yang, Y.~N. Wu, and S.-C. Zhu.
\newblock Interpreting cnns via decision trees.
\newblock {\em In arXiv:1802.00121}, 2018.

\bibitem{InterpretabilitySurvey}
Q.~Zhang and S.-C. Zhu.
\newblock Visual interpretability for deep learning: a survey.
\newblock {\em in Frontiers of Information Technology \& Electronic
  Engineering}, 19(1):27--39, 2018.

\bibitem{CNNSemanticDeep}
B.~Zhou, A.~Khosla, A.~Lapedriza, A.~Oliva, and A.~Torralba.
\newblock Object detectors emerge in deep scene cnns.
\newblock {\em In {ICRL}}, 2015.

\bibitem{CNNSemanticDeep2}
B.~Zhou, A.~Khosla, A.~Lapedriza, A.~Oliva, and A.~Torralba.
\newblock Learning deep features for discriminative localization.
\newblock {\em In {CVPR}}, 2016.

\end{thebibliography}
}

\newpage
\section*{Appendix: Visualization of feature maps of the explainer and feature maps of the performer}

\vspace{20pt}
\includegraphics[width=\linewidth]{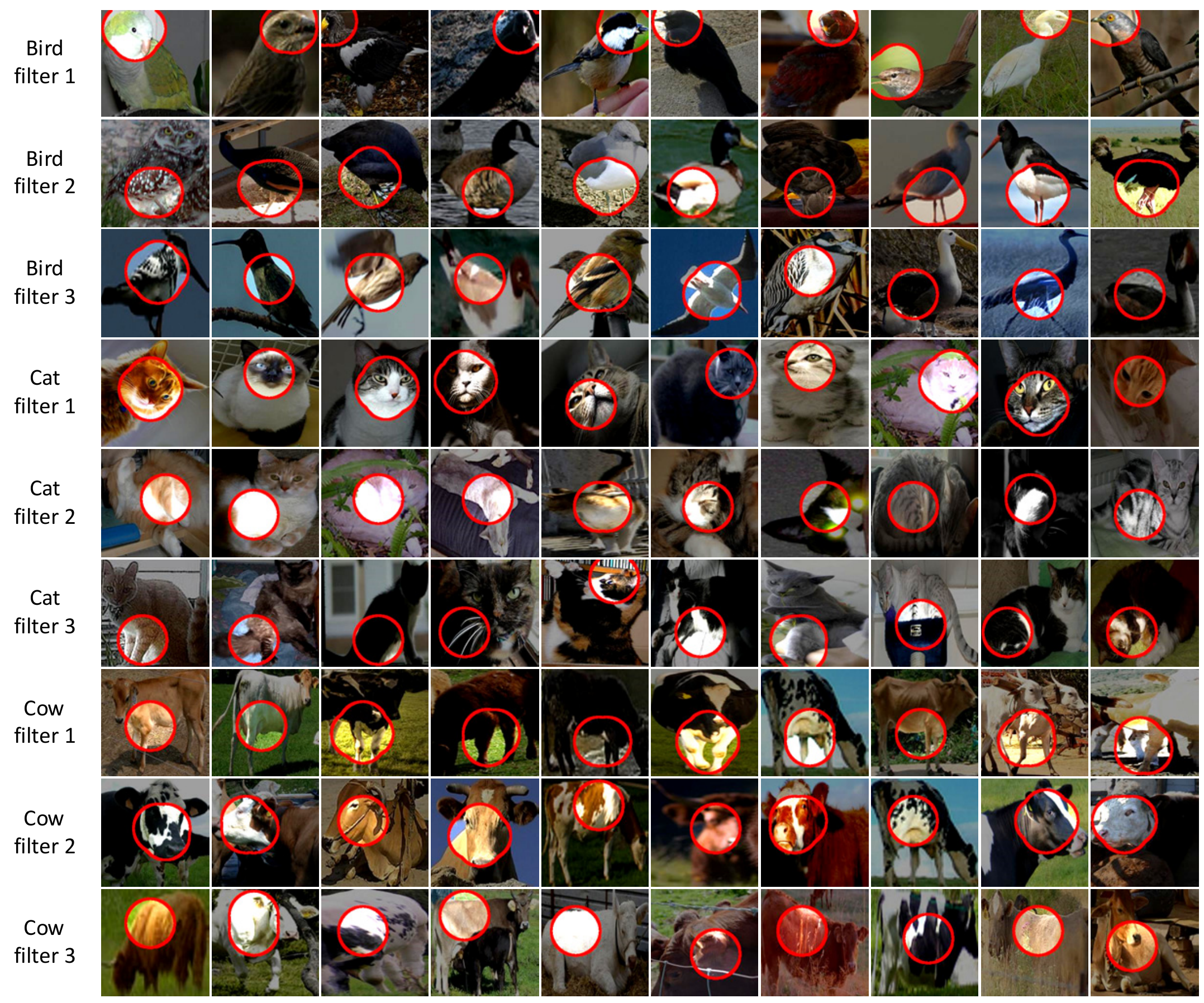}

\vspace{20pt}
Visualization of feature maps in the \textit{conv-interp-2} layer of the explainer. Each row corresponds to feature maps of a filter in the \textit{conv-interp-2} layer. We simply used a round RF for each neural activation and overlapped all RFs for visualization.

\newpage
\vspace{20pt}
\includegraphics[width=\linewidth]{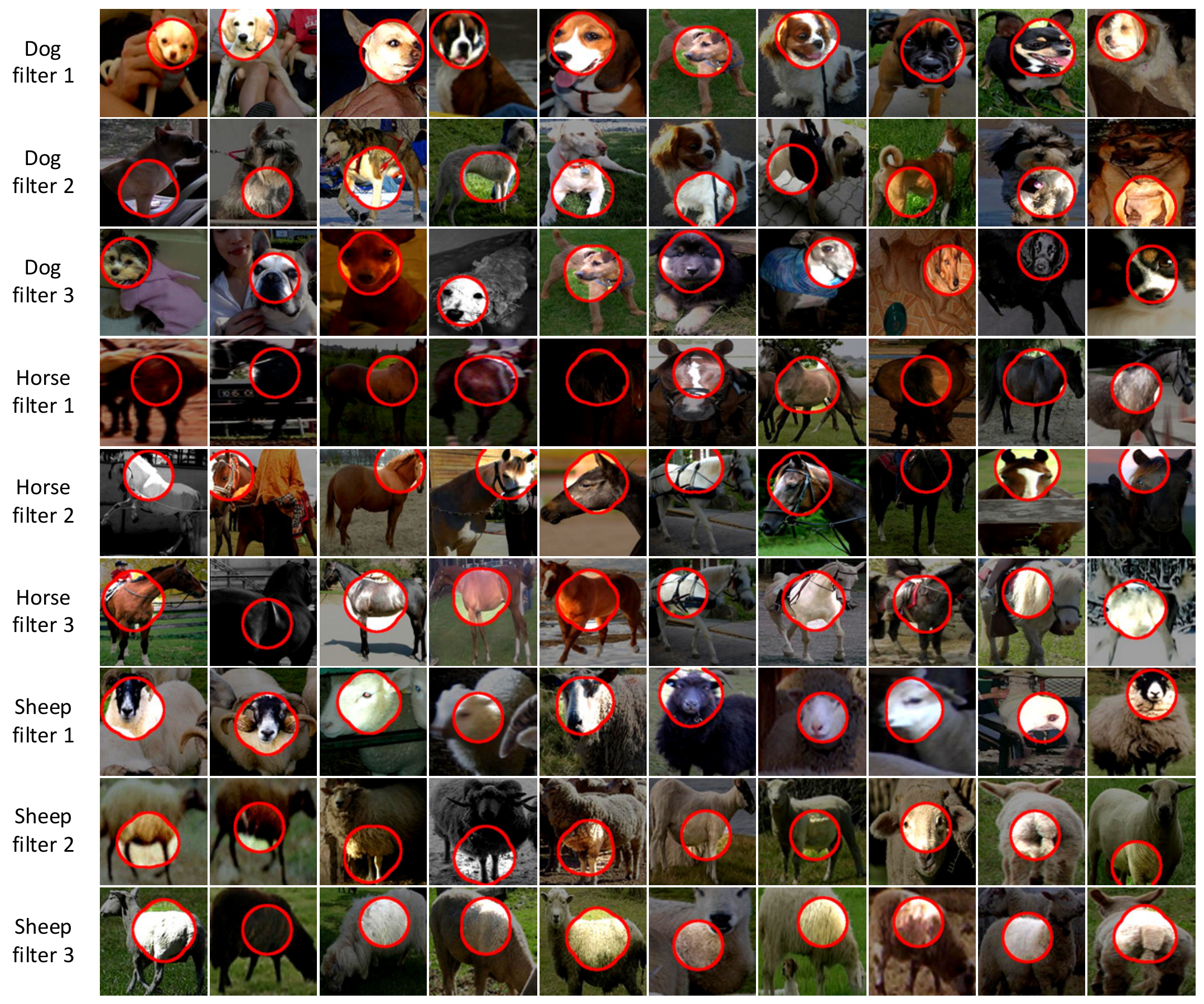}

\vspace{20pt}
Visualization of feature maps in the \textit{conv-interp-2} layer of the explainer. Each row corresponds to feature maps of a filter in the \textit{conv-interp-2} layer. We simply used a round RF for each neural activation and overlapped all RFs for visualization.

\newpage
\includegraphics[width=0.99\linewidth]{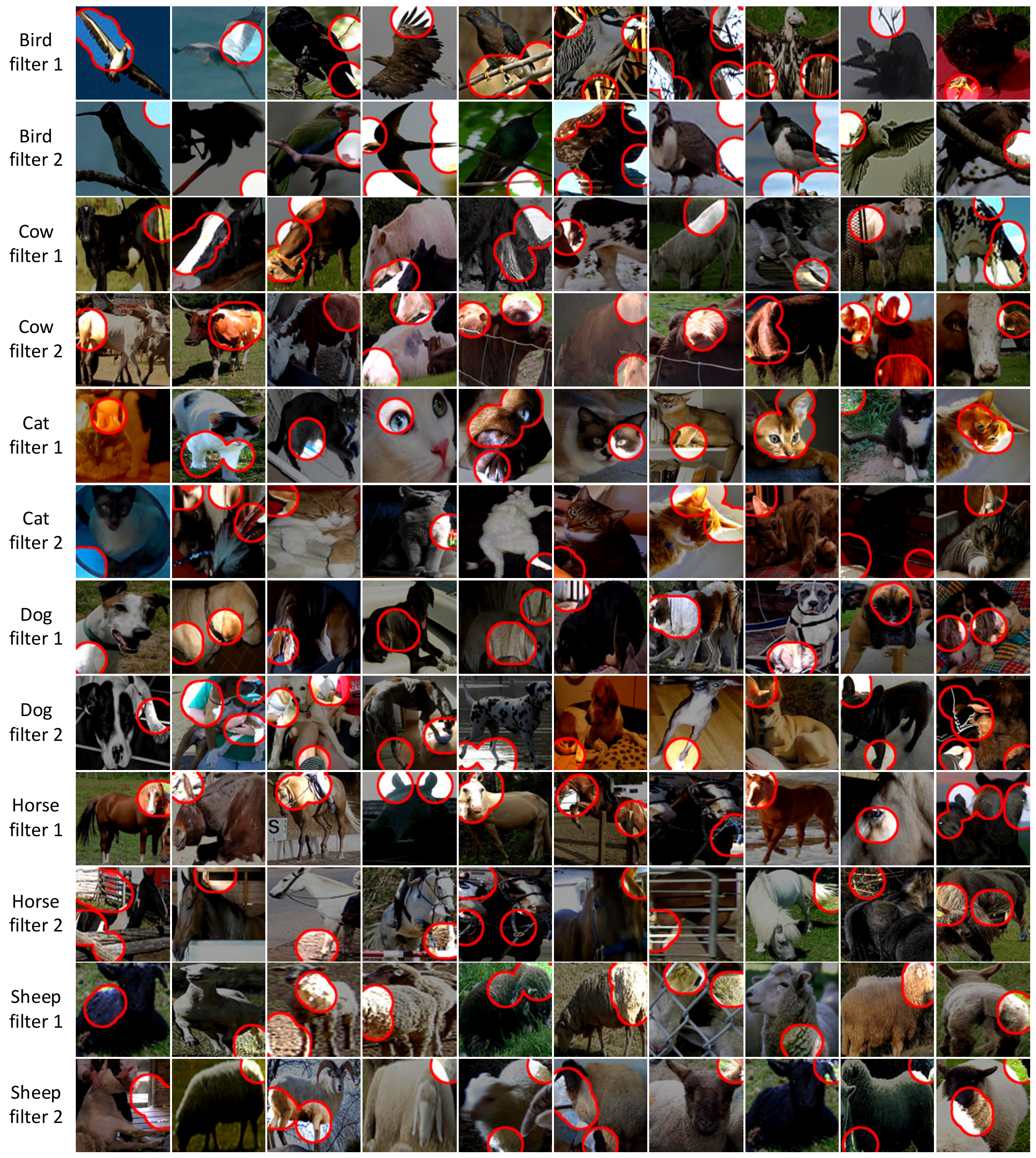}

\vspace{2pt}
Visualization of feature maps in the top conv-layer of the performer. Each row corresponds to feature maps of a filter in the top conv-layer. We simply used a round RF for each neural activation and overlapped all RFs for visualization.

\section*{Appendix: Grad-CAM attention maps}

\includegraphics[width=0.99\linewidth]{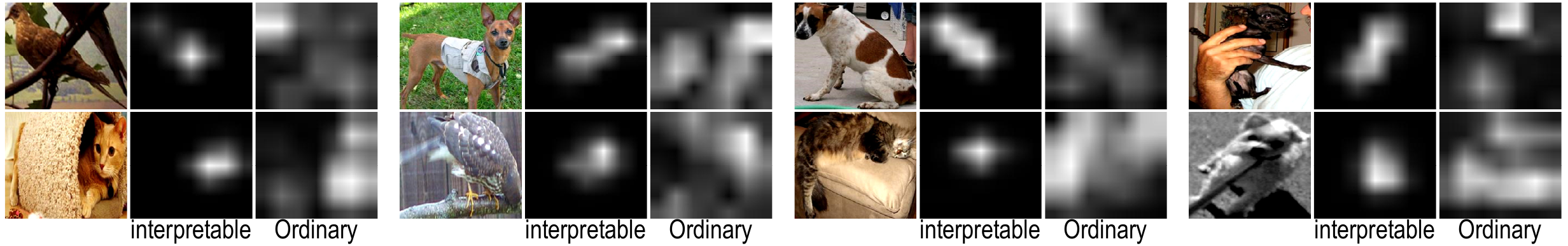}

\vspace{2pt}
Grad-CAM attention maps. We used \cite{visualCNN_grad_2} to compute grad-CAM attention maps of interpretable features of the \textit{conv-interp-2} layer in the explainer and those of ordinary features of the top conv-layer in the performer. Interpretable filters in the \textit{conv-interp-2} layer focused on distinct object parts, while ordinary filters in the performer separated its attention to both textures and parts.

The Quantitative analysis in Figure 5 of the paper shows an example of how to use the disentangled object-part features to quantitatively evaluate contributions of different parts to the output score of object classification. Based on prior semantic meanings of the interpretable filters, we show a prior explanation of the logic in the classification without manually checking activation distributions of each channel of the feature map. Thus, this is different from the visualization of CNN representations, which requires people to manually check the explanation based on visualization results.

\section*{Appendix: Detailed results of $p$ values}

\begin{table}[h]
\caption{\large $p$ values of explainers.}
\resizebox{1.0\linewidth}{!}{\begin{tabular}{l|ccccccc|c|c}
\hline
& \multicolumn{8}{c|}{Pascal-Part dataset~\cite{SemanticPart}} & CUB200-2011~\cite{CUB200}\\
& \multicolumn{7}{c|}{Single-category} & Multi-category & \\
& bird & cat & cow & dog & horse & sheep & Avg. & Avg. & Avg. \\
\hline
{\footnotesize AlexNet} &0.75
&0.72
&0.69
&0.72
&0.70
&0.70
&\textcolor{blue}{0.71}
&--
&0.5810\\
{\footnotesize VGG-M} &0.81
&0.80
&0.81
&0.80
&0.81
&0.81
&\textcolor{blue}{0.81}
&0.9012
&0.8611\\
{\footnotesize VGG-S} &0.91
&0.90
&0.90
&0.90
&0.90
&0.90
&\textcolor{blue}{0.90}
&0.9270
&0.9533\\
{\footnotesize VGG-16} &0.88
&0.89
&0.87
&0.87
&0.86
&0.88
&\textcolor{blue}{0.87}
&0.8593
&0.9579\\
\hline
\end{tabular}}
\end{table}

\section*{Appendix: More results of location instability}

In this section, we add another baseline for comparison. Because we considered feature maps of the \textit{relu4} layer of the AlexNet/VGG-M/VGG-S (the 12th/12th/11th layer of the AlexNet/VGG-M/VGG-S) and the \textit{relu5-2} layer of the VGG-16 (the 28th layer) as target feature maps to be explained, we sent feature maps of these layers feature into explainer networks to disentangle them. Thus, we measured the location instability of these target feature maps as the new baseline.

The following two tables show that our explainer networks successfully disentangled these target feature maps, and the disentangled feature maps in the explainer exhibited much lower location instability.

\begin{table}[h]
\caption{Location instability of feature maps in performers and explainers. Performers are learned based on the Pascal-Part dataset~\cite{SemanticPart}}
\resizebox{0.8\linewidth}{!}{\begin{tabular}{l|ccccccc}
\hline
&\multicolumn{7}{|c}{\large Single-category}\\
\hline
& bird & cat & cow & dog & horse & sheep & Avg\\
\hline
AlexNet (the \textit{relu4} layer) &0.152
&0.130
&0.140
&0.127
&0.143
&0.139
&\textcolor{blue}{0.139}\\
AlexNet (the top conv-layer) &0.153
&0.131
&0.141
&0.128
&0.145
&0.140
&\textcolor{blue}{0.140}\\
Explainer &{\bf0.104}
&{\bf0.089}
&{\bf0.101}
&{\bf0.083}
&{\bf0.098}
&{\bf0.103}
&\textcolor{blue}{\bf0.096}\\
\hline
VGG-M (the \textit{relu4} layer) &0.148
&0.127
&0.138
&0.126
&0.140
&0.137
&\textcolor{blue}{0.136}\\
VGG-M (the top conv-layer) &0.152
&0.132
&0.143
&0.130
&0.145
&0.141
&\textcolor{blue}{0.141}\\
Explainer &{\bf0.106}
&{\bf0.088}
&{\bf0.101}
&{\bf0.088}
&{\bf0.097}
&{\bf0.101}
&\textcolor{blue}{\bf0.097}\\
\hline
VGG-S (the \textit{relu4} layer) &0.148
&0.127
&0.136
&0.125
&0.139
&0.137
&\textcolor{blue}{0.135}\\
VGG-S (the top conv-layer) &0.152
&0.131
&0.141
&0.128
&0.144
&0.141
&\textcolor{blue}{0.139}\\
Explainer &{\bf0.110}
&{\bf0.085}
&{\bf0.098}
&{\bf0.085}
&{\bf0.091}
&{\bf0.096}
&\textcolor{blue}{\bf0.094}\\
\hline
VGG-16 (the \textit{relu5-2} layer) &0.151
&0.128
&0.145
&0.124
&0.146
&0.146
&\textcolor{blue}{0.140}\\
VGG-16 (the top conv-layer) &0.145
&0.133
&0.146
&0.127
&0.143
&0.143
&\textcolor{blue}{0.139}\\
Explainer &{\bf0.095}
&{\bf0.089}
&{\bf0.097}
&{\bf0.085}
&{\bf0.087}
&{\bf0.089}
&\textcolor{blue}{\bf0.090}\\
\hline
\end{tabular}}
\end{table}

\begin{table}[h]
\caption{Location instability of feature maps in performers and explainers. Performers are learned based on the CUB200-2011 dataset~\cite{CUB200}}
\begin{tabular}{lc}
\hline
AlexNet (\textit{relu4} layer) & 0.1542\\
AlexNet (the top conv-layer) & 0.1502\\
Explainer & {\bf0.0906}\\
\hline
VGG-M (\textit{relu4} layer) & 0.1484\\
VGG-M (the top conv-layer) & 0.1476\\
Explainer & {\bf0.0815}\\
\hline
VGG-S (\textit{relu4} layer) & 0.1518\\
VGG-S (the top conv-layer) & 0.1481\\
Explainer & {\bf0.0704}\\
\hline
VGG-16 (\textit{relu5-2} layer) &0.1444\\
VGG-16 (the top conv-layer) & 0.1373\\
Explainer& {\bf0.0490}\\
\hline
\end{tabular}
\end{table}

\section*{Appendix: Understanding of filter losses}

We can re-write the filter loss as
\begin{equation}
{\bf Loss}_{f}=-H({\bf T})+H({\bf T}'|{\bf X})+{\sum}_{x_{f}\in{\bf X}}p({\bf T}^{+},x_{f})H({\bf T}^{+}|X=x_{f})\nonumber
\end{equation}
where ${\bf T}'=\{T^{-},{\bf T}^{+}\}$. $H({\bf T})=-\sum_{T\in{\bf T}}p(T)\log p(T)$ is a constant prior entropy of part templates.

Low inter-category entropy: The second term $H({\bf T}'=\{T^{-},{\bf T}^{+}\}|{\bf X})$ is computed as $H({\bf T}'=\{T^{-},{\bf T}^{+}\}|{\bf X})=-\sum_{x_{f}}p(x_{f})\sum_{T\in\{T^{-},{\bf T}^{+}\}}p(T|x_{f})\log p(T|x_{f})$, where ${\bf T}^{+}=\{T_{\mu_1},T_{\mu_2},\ldots,T_{\mu_{L^2}}\}\subset{\bf T}$ and $p({\bf T}^{+}|x_{f})=\sum_{\mu}p(T_{\mu}|x_{f})$. We define the set of all positive templates {${\bf T}^{+}$} as a single label to represent category $c$. We use the negative template {$T^{-}$} to denote other categories. This term encourages a low conditional entropy of inter-category activations, \emph{i.e.} a well-learned filter $f$ needs to be exclusively activated by a certain category $c$ and keep silent on other categories. We can use a feature map $x_{f}$ of $f$ to identify whether the input image belongs to category $c$ or not, \emph{i.e.} $x_{f}$ fitting to either $T_{\hat{\mu}}$ or $T^{-}$, without great uncertainty.

Low spatial entropy: The third term is given as $H({\bf T}^{+}|X=x_{f})=\sum_{\mu}\tilde{p}(T_{\mu}|x_{f})\log\tilde{p}(T_{\mu}|x_{f})$, where $\tilde{p}(T_{\mu}|x_{f})=\frac{p(T_{\mu}|x_{f})}{p({\bf T}^{+}|x_{f})}$. This term encourages a low conditional entropy of spatial distribution of $x_{f}$'s activations. \emph{I.e.} given an image {$I\in{\bf I}_{c}$}, a well-learned filter should only be activated by a single region $\hat{\mu}$ of the feature map $x_{f}$, instead of being repetitively triggered at different locations.

\textbf{Optimization of filter losses:} The computation of gradients of the filter loss \emph{w.r.t.} each element $x_{f}^{(ij)}$ of feature map $x_{f}$ is time-consuming. \cite{interpretableCNN} computes an approximate but efficient gradients to speed up the computation, as follows.
\begin{small}
\begin{equation}
\frac{\partial{\bf Loss}_{f}}{\partial x_{f}^{(ij)}}=\sum_{T}\frac{p(T)t_{ij}e^{tr(x_{f}\cdot T)}}{Z_{T}}\Big\{tr(x_{f}\cdot T)-\log\big[Z_{T}p(x_{f})\big]\Big\}\approx\frac{p(\hat{T})\hat{t}_{ij}}{Z_{\hat{T}}}e^{tr(x_{f}\cdot\hat{T})}\Big\{tr(x_{f}\cdot\hat{T})-\log[Z_{\hat{T}}p(x_{f})]\Big\}\nonumber
\end{equation}
\end{small}
where {$\hat{T}$} is the target template for feature map $x_{f}$. Let us assume that there are multiple object categories $C$. We simply assign each filter $f$ with the category $\hat{c}\in C$ whose images activate $f$ the most, \emph{i.e.} {$\hat{c}={\arg\!\max}_{c}\mathbb{E}_{x_{f}:I\in{\bf I}_{c}}\sum_{ij}x_{f}^{(ij)}$}. If the input image $I$ belongs to the target category of filter $f$, then $\hat{T}=T_{\hat{\mu}}$, where {$\hat{\mu}\!=\!{\arg\!\max}_{\mu\!=\![i,j]}x_{f}^{(ij)}$}. If image $I$ belongs to other categories, then {$\hat{T}\!=\!T^{-}$}. Considering {$\forall T\!\in\!{\bf T}\setminus\{\hat{T}\}$}, {$e^{tr(x_{f}\cdot\hat{T})}\!\gg\!e^{tr(x_{f}\cdot T)}$} after initial learning episodes, we can make approximations in the above equation.

Note that above assignments of object categories are also used to compute location instability for middle-layer filters to evaluate their interpretability.

Inspired by optimization tricks in \cite{interpretableCNN}, we updated the parameter {$\lambda_{f}=\frac{1}{300N}\mathbb{E}_{x_{f}}[\Vert\frac{\partial Loss_{rec}}{\partial x_{f}}\Vert]/\mathbb{E}_{x_{f}}[\Vert\frac{\partial Loss_{f}}{\partial x_{f}}\Vert]$} for the $N$-th learning epoch in an online manner, where {$\frac{\partial Loss_{rec}}{\partial x_{f}}$} denotes gradients of reconstruction losses obtained from upper layers. In particular, given performers for single-category classification, we simply used feature maps of positive images (\emph{i.e.} objects of the target category) to approximately estimate the parameter ${\boldsymbol\alpha}$ for the norm-layer, because positive images can much more strongly trigger interpretable filters than negative images. Thus, computing ${\boldsymbol\alpha}$ based on positive images made $p$ accurately measure the contribution ratio of the interpretable track when the network made predictions to positive images. We will clarify all these settings when the paper is accepted. In experiments, for interpretable filters in the \textit{conv-interp-2} layer, we added the filter loss to {$x'_{f}=p\cdot x_{f}+(1-p)\cdot\hat{x}_{\textrm{ordin}}$}, where {$\hat{x}_{\textrm{ordin}}\in\mathbb{R}^{L\times L}$} denotes a channel of $x_{\textrm{ordin}}$ that corresponds to the channel of filter $f$. We found that this modification achieved more robust performance than directly applying the filter loss to $x_{f}$. In this case, the filter loss encouraged a large value of $p$ and trained the interpretable filter $f$, but we did not pass gradients of the filter loss to the ordinary track.

\section*{Appendix: Evaluating the reconstruction quality based on the object-classification accuracy}

\begin{table}[h]
\resizebox{\linewidth}{!}{\begin{tabular}{l|ccc|ccc|ccc}
\hline
& \multicolumn{6}{c|}{Pascal-Part~\cite{SemanticPart}} & \multicolumn{3}{|c}{CUB200~\cite{CUB200}}\\
& \multicolumn{3}{c|}{Multi-category} & \multicolumn{3}{c|}{Single-category} &\\
\!\!\!&\!\!\! {\scriptsize Performer} \!\!\!&\!\!\! {\scriptsize Explainer} \!\!\!&\!\!\! {\scriptsize Explainer+cls}
\!\!\!&\!\!\! {\scriptsize Performer} \!\!\!&\!\!\! {\scriptsize Explainer} \!\!\!&\!\!\! {\scriptsize Explainer+cls}
\!\!\!&\!\!\! {\scriptsize Performer} \!\!\!&\!\!\! {\scriptsize Explainer} \!\!\!&\!\!\! {\scriptsize Explainer+cls}\!\!\!\\
\hline
\!\!\!AlexNet \!\!\!&\!\!\! -- \!\!\!&\!\!\! -- \!\!\!&\!\!\! --
\!\!\!&\!\!\! 4.60\% \!\!\!&\!\!\! 8.20\% \!\!\!&\!\!\! 2.88\%
\!\!\!&\!\!\! 4.41\% \!\!\!&\!\!\! 10.98\% \!\!\!&\!\!\! 3.57\%
\!\!\!\\
\!\!\!VGG-M \!\!\!&\!\!\! 6.12\% \!\!\!&\!\!\! 6.62\% \!\!\!&\!\!\! 5.22\%
\!\!\!&\!\!\! 3.18\% \!\!\!&\!\!\! 8.58\% \!\!\!&\!\!\! 3.40\%
\!\!\!&\!\!\! 2.66\% \!\!\!&\!\!\! 6.84\% \!\!\!&\!\!\! 2.54\%
\!\!\!\\
\!\!\!VGG-S \!\!\!&\!\!\! 5.95\% \!\!\!&\!\!\! 6.97\% \!\!\!&\!\!\! 5.43\%
\!\!\!&\!\!\! 2.26\% \!\!\!&\!\!\! 10.97\% \!\!\!&\!\!\! 3.86\%
\!\!\!&\!\!\! 2.76\% \!\!\!&\!\!\! 8.53\% \!\!\!&\!\!\! 2.72\%
\!\!\!\\
\!\!\!VGG-16 \!\!\!&\!\!\! 2.03\% \!\!\!&\!\!\! 2.17\% \!\!\!&\!\!\! 2.49\%
\!\!\!&\!\!\! 1.34\% \!\!\!&\!\!\! 6.12\% \!\!\!&\!\!\! 1.76\%
\!\!\!&\!\!\! 1.09\% \!\!\!&\!\!\! 6.04\% \!\!\!&\!\!\! 0.90\%
\!\!\!\\
\hline
\end{tabular}}
\caption{\large Classification errors based on feature maps of performers and explainers.}
\end{table}

In order to evaluate the feature-reconstruction quality, we used the classification accuracy based on explainer features as an evaluation metric. We fed output features of the explainer back to the performer for classification. Theoretically, a high classification accuracy may demonstrate that the explainer can well reconstruct performer features without losing much information. Note that explainers were learned to reconstruct feature maps of performers, rather than optimizing the classification loss, so explainers could only approximate the classification performance of performers but could not outperform performers.

In addition, we added another baseline, namely \textit{Explainer+cls}, which used the object-classification loss to replace the reconstruction loss to learned explainer networks. Thus, output features of \textit{Explainer+cls} exhibited higher classification accuracy than features of the original explainer.

The following table compares the classification accuracy between the performer and the explainer. For multi-category classification, the performance of explainers was quite close to that of performers. Learning explainers with classification losses exhibited significantly better classification performance than learning explainers with reconstruction losses. Because \textit{Explainer+cls} directly learned from the classification loss, \textit{Explainer+cls} sometimes even outperformed the performer.

\end{document}